\documentclass[a4paper, 10pt, conference]{ieeeconf}      
\usepackage{FG2026}
\usepackage{multirow}
\usepackage{graphicx}
\usepackage{booktabs}
\usepackage{subcaption}
\usepackage{hyperref}
\FGfinalcopy 

\IEEEoverridecommandlockouts                              
\def\FGPaperID{342} 

\title{\LARGE \bf
Employing Vision-Language Models \\ for Face Image Quality Assessment
}

\author{\parbox{16cm}{\centering
    {\large Erdi Sarıtaş$^1$, Eren Onaran$^1$, Vitomir Štruc$^2$, and Hazım Kemal Ekenel$^{1,3}$}\\
    {\normalsize
    $^1$ Department of Computer Engineering, Istanbul Technical University, Istanbul, Türkiye\\
    $^2$ Faculty of Electrical Engineering, University of Ljubljana, Ljubljana, Slovenia\\
    $^3$ Division of Engineering, NYU Abu Dhabi, Abu Dhabi, UAE}}
    \thanks{This work was supported by the Meetween Project that received funding from the European Union's Horizon Europe Research and Innovation Programme under Grant Agreement No. 101135798 and supported in parts by the ARIS grant P2-0250. Computing resources used in this work were provided by the National Center for High Performance Computing of Turkey (UHeM) under grant number 4023702025.
}
}

\usepackage{fancyhdr}
\begin{document}

\ifFGfinal
\thispagestyle{empty}
\pagestyle{empty}
\else
\author{\parbox{16cm}{\centering
    {\large Erdi Sarıtaş$^1$, Eren Onaran$^1$, Vitomir Štruc$^2$, and Hazım Kemal Ekenel$^{1,3}$}\\
    {\normalsize
    $^1$ Department of Computer Engineering, Istanbul Technical University, Istanbul, Türkiye\\
    $^2$ Faculty of Electrical Engineering, University of Ljubljana, Ljubljana, Slovenia\\
    $^3$ Division of Engineering, NYU Abu Dhabi, Abu Dhabi, UAE}}
    \thanks{This work was supported by the Meetween Project that received funding from the European Union's Horizon Europe Research and Innovation Programme under Grant Agreement No. 101135798 and supported in parts by the ARIS grant P2-0250. Computing resources used in this work were provided by the National Center for High Performance Computing of Turkey (UHeM) under grant number 4023702025.
}
}
\pagestyle{plain}
\fi
\maketitle
\thispagestyle{fancy}

\begin{abstract}

Face Image Quality Assessment (FIQA) is a crucial control step in biometric pipelines. It ensures only reliable samples are processed to maintain system accuracy. State-of-the-art FIQA methods achieve high utility but typically operate as "black boxes." They produce scalar scores without human-interpretable justifications. This lack of transparency limits their effectiveness in human-in-the-loop scenarios, such as automated border control, where actionable feedback is essential. In this paper, we investigate the potential of off-the-shelf Vision-Language Models (VLMs) to bridge this gap by performing FIQA in a zero-shot setting. 
We present a comprehensive evaluation framework for assessing VLM performance. This involves benchmarking traditional FIQA methods through error-versus-reject curves. Additionally, using a diverse set of datasets, ranging from surveillance-oriented to synthetically generated, we analyzed their interpretability, consistency, and robustness to prompt changes.
Our results show biometric utility performance depends significantly on architecture, not merely on parameter count. Most VLMs' outputs align with those of traditional methods. We also find that VLM ranking performance and the generated scores may vary across prompts. Our synthetic ablation study shows that while increasing the parameter count can improve internal consistency, it yields worse degradation-detection performance than smaller models.
These findings suggest that zero-shot FIQA score estimation using VLMs is promising and could effectively complement conventional FIQA pipelines as an interpretability module. The codes are available at \href{https://github.com/ThEnded32/VLM4FIQA.git}{github.com/ThEnded32/VLM4FIQA.git}.

\end{abstract}

\section{Introduction}
\label{sec:intro}

\begin{figure}[!t]
  \centering
  \includegraphics[width=0.95\linewidth]{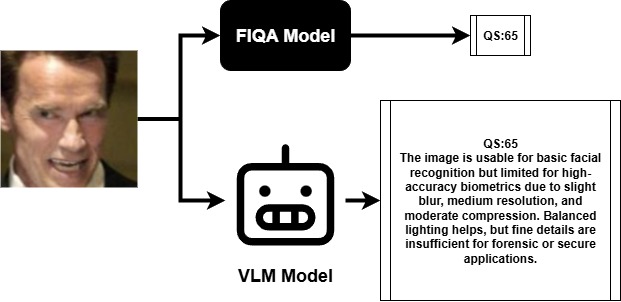}
  \caption{\textbf{VLMs for Quality Assessment.} While traditional FIQA methods (top) function as opaque "black boxes" outputting only scalar scores, VLM-driven approaches (bottom) offer transparency by providing both biometric utility scores and actionable semantic justifications.\\
  $^{*}$\footnotesize{VLM prompt is generated using QWEN2.5-32B.}
  }
  \label{fig:intro}
\end{figure}

Face Image Quality Assessment (FIQA) serves as a critical control step in biometric systems, ensuring that only reliable samples are processed.
Low-quality images, whether caused by blur, poor lighting, or other factors, can significantly increase false-rejection rates, which poses a risk to system consistency.
Consequently, FIQA has become a mandatory pre-processing step in modern face recognition pipelines~\cite{schlett2022face}.
State-of-the-art (SOTA) FIQA methods have primarily benefited from advances in deep learning, relying on latent quality signals directly from face recognition models.
Approaches based on uncertainty estimation~\cite{terhorst2020ser}, margin-based learning~\cite{kim2022adaface}, and diffusion-based reconstruction~\cite{babnik2024ediffiqa} have achieved impressive correlation with biometric utility.
However, they operate under a "black-box" paradigm: they produce a single scalar score without providing interpretable explanations.
This lack of transparency leads to the absence of actionable feedback, e.g., "please come closer", in human-oriented scenarios, like automated border control.
Fig.~\ref{fig:intro} illustrates this limitation, contrasting the opaque scalar output of traditional methods with the transparent, descriptive feedback offered by the VLM-driven paradigm.
Providing this human-understandable feedback can enhance both interaction and capture quality.

Concurrently, the literature is witnessing a paradigm shift driven by Vision-Language Models (VLMs).
In the domain of general Image Quality Assessment (IQA), research has moved beyond simple regression, utilizing VLMs to generate descriptive quality evaluations and reason about aesthetic or technical attributes~\cite{wu2024qalign,you2025edqa}.
Similarly, in the domain of biometrics, VLMs are increasingly employed to provide semantic explanations for tasks such as attribute recognition and face anti-spoofing~\cite{chaubey2025facellava,lin2025instructflip}.
This convergence suggests that multimodal models have the intrinsic capacity to bridge the interpretability gap in FIQA.
However, it remains an open question whether the general visual reasoning of these models aligns with the utility-centric needs of biometric quality assessment.

This work investigates the fundamental research question: \emph{Can the zero-shot capacity of general-purpose VLMs align with the requirements of biometric quality assessment?} To address this, we structure our study around four inquiries:

\begin{itemize}
    \item \textbf{Biometric Utility:} Can off-the-shelf VLMs achieve error-versus-reject performance comparable to specialized, supervised FIQA methods?
    \item \textbf{Interpretability:} Do the generated semantic descriptions accurately reflect physical degradations in real-world surveillance scenarios?
    \item \textbf{Consistency vs. Precision:} How does model scale influence the trade-off between hallucination rates on clean data and the internal consistency of quality scores?
    \item \textbf{Prompt Robustness:} Is the zero-shot ranking logic stable across different semantic prompt phrasings, or does it require specific triggers to function correctly?
\end{itemize}

To answer these questions, we introduce a comprehensive evaluation framework that prioritizes assessing biometric utility using error-versus-reject (EvR) curves. In addition to this primary analysis, we investigate the models' sensitivity to real-world degradation in surveillance scenarios. We also examine the cross-prompt stability of their scoring logic and their interpretability across diverse datasets. Furthermore, we conduct a controlled synthetic ablation study to rigorously assess the precision of their degradation-detection capabilities. Lastly, we present example VLM outputs to illustrate their image-specific responses.

Our study shows that architectural choices determine zero-shot capability, rather than solely parameter count. Some general-purpose models have been found to be competitive with traditional FIQA estimators in terms of biometric scoring performance. Crucially, analysis of our synthetic dataset ablation reveals a potential complex trade-off between utility and descriptive precision: while larger models maintain robust rankings and strict internal consistency, they exhibit significantly higher hallucination rates on clean images than their smaller counterparts. Furthermore, we find that vulnerability to prompt phrasing is relatively scale-dependent: smaller models are more influenced by specific semantic keywords, whereas larger architectures exhibit a more stable ranking logic that is invariant to phrasing.

\section{Related Work}
\label{sec:related}

\subsection{Learning-based FIQA Methods}
Learning-based FIQA approaches primarily rely on features from face recognition (FR) embeddings. SER-FIQ~\cite{terhorst2020ser} measures quality based on embedding stability under stochastic dropout, and AdaFace~\cite{kim2022adaface} uses the feature vector's magnitude as a quality proxy during training. Further methods focus on intrinsic FR properties; SDD-FIQA~\cite{ou2021sddfiqa} is an unsupervised approach that uses the similarity distribution distance between intra- and inter-class samples, and FaceQAN~\cite{babnik2022faceqan} links quality to adversarial robustness by considering the noise magnitude. More recently, advanced architectures have emerged. Generative approaches, such as eDifFIQA~\cite{babnik2024ediffiqa}, correlate quality with reconstructive performance using diffusion models. Vision Transformer (ViT)-based methods are prominent, including ViT-FIQA~\cite{atzori2025vitfiqa}, which uses a learnable "quality token", and DSL-FIQA~\cite{chen2024dslfiqa}, which employs landmark-guided transformers with dual-set degradation learning. Finally, MR-FIQA~\cite{ou2025mrfiqa} addresses data scarcity by leveraging synthetic data and multi-reference representations. Despite high utility in biometrics tasks such as FR, these methods only output scalar scores, lacking human-interpretable justifications.

\subsection{VLM-based Image Quality Assessment}
Vision-Language Models (VLMs) have fundamentally transformed general image quality assessment by enabling human-interpretable reasoning. Early works employed CLIP~\cite{radford2021learning} embeddings for zero-shot assessment via prompting comparative judgments (e.g., "sharp" versus "noisy")~\cite{wang2023exploring,zhang2023blind,miyata2024zen}. The field progressed to generative scoring and reasoning. Generative methods such as Q-Align~\cite{wu2024qalign} train LMMs to predict discrete quality tokens (e.g., "Good"). Additionally, a paradigm refined by the team "Next" in the VQualA 2025 Challenge~\cite{ma2025vquala} adapts the CLIP model to predict the image quality distribution via a multi-level quality-aware prompt learning mechanism. Other models focus on reasoning using VLMs, such as Co-Instruct~\cite{wu2024openended} for multi-image comparison and EDQA~\cite{you2025edqa} for generating descriptive quality paragraphs. QA-VLM~\cite{zheng2025qavlm} applied VLMs to domain-specific quality (additive manufacturing) for interpretable reasoning. However, these generalist models are not explicitly optimized for biometric face utility.

\subsection{VLM-based Face Analysis and Explainability}
VLMs have also emerged for face analysis tasks requiring deep semantic understanding. These models generate human-readable explanations for diverse human-centric tasks. For identity and attribute analysis, FaceLLM~\cite{otroshi2025facellm} and Face-LLaVA~\cite{chaubey2025facellava} generate explanations for identity verification and facial attributes. Broader benchmarks, such as HERM~\cite{zhang2024herm}, assess VLMs across diverse human understanding tasks. Furthermore, VLMs have been adapted to the security and forgery-detection domain. MGFFD-VLM~\cite{chen2025mgffdvlam} is a deepfake detector using multi-granularity prompts to reason about quality-related artifacts and generate rationales. Other security works include FaceShield~\cite{wang2025faceshield}, which provides reasoning for anti-spoofing decisions, and InstructFLIP~\cite{lin2025instructflip}, which uses unified instruction tuning to generalize across presentation attacks. Collectively, these works demonstrate that VLMs can be effectively utilized for facial semantics understanding and security-critical control.

\subsection{Explainable and VLM-driven Face Quality Assessment}

Explainable Face Image Quality Assessment (X-FIQA) aims to enhance the interpretability of FIQA models by providing human-understandable justifications through visual and semantic outputs. Visuals such as heatmaps can be generated using intrinsic properties of methods, e.g., landmark-guided attention of DSL-FIQA~\cite{chen2024dslfiqa} to produce an attention heatmap. Furthermore, IFQA~\cite{jo2023ifqa} converts the real/fake paradigm of GANs to high/low quality, and trains a discriminator that produces pixel-level scores. With this, they can directly generate a quality heatmap by default. On the semantic interpretability side, Face Quality Vector (FQV)~\cite{najafzadeh2023face} introduces a multidimensional quality representation that captures expression, pose, and illumination, beyond a single scalar quality value.

Most recently, researchers have begun bridging the gap between FIQA and VLMs. CLIB-FIQA~\cite{ou2024clibfiqa} introduced confidence calibration using CLIP to anchor objective quality factors like blur and occlusion. Moving to generative assessment, FVQ-Rater~\cite{wu2025fvq} utilizes instruction tuning to score face video quality, and MDTFIQA~\cite{gao2025mdfiqa} evaluates the fidelity of text-to-face generation. Finally, FaceOracle~\cite{kabbani2025faceoracle} focuses on interactive transparency, using Retrieval-Augmented Generation~\cite{lewis2020retrieval} (RAG), and can be used to explain quality scores via a chat interface. Our research advances this emerging field by providing a comprehensive quantitative analysis of VLM usage in a zero-shot setting.

\section{Proposed Method}
\label{sec:method}

We formulate a general evaluation framework to assess the performance of Vision-Language Models (VLMs) on the FIQA task in a zero-shot setting. This section introduces the datasets, the prompt types used to query the models, and the evaluation pipeline, which prioritizes biometric utility analysis.

\subsection{Datasets}
We use diverse real-world face datasets to verify performance across different capture conditions:
\begin{itemize}
    \item \textbf{CelebA-HQ}~\cite{karras2018progressive}: Provides high-quality, studio-like aligned faces, serving as a reference distribution for "good" quality.
    \item \textbf{LFW}~\cite{huang2007labeled} and \textbf{IJB-B}~\cite{whitelam2017ijbb}: Contain in-the-wild face images with natural variations in pose, illumination, and occlusion. These are standard benchmarks for assessing biometric utility under realistic variability.
    \item \textbf{SCFace}~\cite{grgic2011scface}: A surveillance dataset capturing subjects at three standoff distances (4.2~m, 2.6~m, and 1.0~m). We use this to analyze sensitivity to physical degradation (resolution and blur) in a controlled security scenario.
\end{itemize}
In all cases, faces are detected and aligned using MTCNN~\cite{zhang2016joint} and resized to a fixed resolution of $224 \times 224$.

\subsection{Prompt Design}
We query the target VLMs using two primary strategies: scalar scoring for utility analysis and attribute classification for explainability.

\textbf{Simple Quality Prompt:} The primary prompt asks the model to rate a face image on a scale from 0 to 100. The model acts as an \textit{"expert image quality assessor for face images"} and is prompted to \textit{"evaluate the image quality for facial analysis"}. The model generates a strict JSON object containing a single numeric field (i.e., \textit{\{"Quality Score": $<$0-100$>$\}}), and no other text or keys. 

\textbf{Attribute Classification Prompt:} For assessing explainability, we use a structured prompt that requests categorical judgments on specific quality factors, with degradation level options. The model returns a JSON object containing a scalar quality score alongside attributes defined by strict options:

\begin{itemize}
    \item \textit{"Sharpness"}: Clear, Slightly-, Moderately-, or Strongly Blurred.
    \item \textit{"Resolution"}: High, Medium, Low, or Very Low.
    \item \textit{"Lighting"}: Balanced, plus three intensities for Dark and Bright.
    \item \textit{"Compression"}: None, Minimal, Moderate, or Severe.
\end{itemize}
This allows us to verify internal consistency by checking if the predicted score aligns with the detected artifact severity.

\textbf{Semantic Variants:} We test two additional semantic phrasings against the simple baseline to investigate the behavior of different scalar scoring prompts regarding biometric utility. For these variants, we retain the expert role definition but update the specific instruction and the required JSON output key:
\begin{itemize}
    \item \textit{Utility}: Prompted to \textit{"evaluate the image utility for a face recognition model"} using the key \textit{"Utility Score"}.
    \item \textit{Reliability}: Prompted to \textit{"evaluate the image reliability for face verification"} using the key \textit{"Reliability Score"}.
\end{itemize}

\subsection{Evaluation Pipeline and Metrics}
Our evaluation has three stages, shifting the focus from simple correlation to decision-based biometric performance.

\paragraph{Stage 1: Biometric Utility (Error-versus-Reject)}
The core of our benchmark is the error-versus-reject (EvR) analysis. A reliable FIQA estimator should assign lower scores to images that cause recognition errors.
\begin{itemize}
    \item \textbf{Protocol}: We compute the False Non-Match Rate (FNMR) at fixed False Match Rate (FMR) thresholds (i.e., $10^{-3}$). We progressively reject a fraction of the dataset with the lowest predicted quality scores.
    \item \textbf{Metrics}: We report the \textbf{Area Under the Curve (AUC)} of the EvR plot. A lower AUC indicates that the quality scores successfully prioritize reliable images. We also report \textbf{partial AUC} at specific low rejection ratios (1\%, 5\%, 10\%, and 20\%) to measure effectiveness in strictly operational ranges where high data retention is required:
    \begin{equation}
        \label{eq:pauc}
        pAUC_\rho = \frac{1}{\rho} \int_{0}^{\rho} \mathrm{FNMR}(r) \, dr
    \end{equation}
\end{itemize}

\paragraph{Stage 2: Surveillance Sensitivity}
Using SCFace, we analyze whether the VLM assigns higher quality scores to images captured at closer distances (1.0~m) than to those captured at farther distances (4.2~m). We further analyze the Attribute Classification outputs to determine whether the model explicitly assigns lower scores to relevant factors such as "Low Resolution."

\paragraph{Stage 3: Consistency}
We assess the stability of the model's scoring logic by measuring:
\begin{itemize}
    \item \textbf{Cross-Prompt Consistency}: The Mean Absolute Error (MAE), Pearson correlation, and bias between scores generated by the Quality (both Simple and Classification), Utility, and Reliability prompts.
    \item \textbf{Internal Consistency}: The alignment between scalar scores and categorical labels, e.g., ensuring "Blurry" images receive lower scores than "Clear" ones.
\end{itemize}

\subsection{Synthetic Ablation Study: Mix Degradation}
\label{sec:method_synthetic}

To rigorously assess the interpretability and sensitivity of VLMs, we design a controlled synthetic ablation study. While real-world datasets provide biometric realism, they lack granular ground truth for specific degradations (e.g., exact noise levels or blur kernels). To address this, we generate a Mix Degradation Benchmark using high-quality samples from CelebA-HQ~\cite{karras2018progressive} as the clean reference baseline.

\paragraph{Degradation Logic}
We define a degradation space $\mathcal{D}$ consisting of five artifact types: blur, noise, low resolution, JPEG compression, and under-/over-exposure. For each clean reference image, inspired by the impact of combined degradations~\cite{saritacs2024analyzing}, we generate a Mixed Degradation variant by applying a combination of three random artifacts: one at high intensity (the "hard artifact") and the other two at mild intensity.

\paragraph{Ground Truth Vectors}
Each image is associated with a binary ground truth vector $v_{GT} \in \{0,1\}^5$, where the $i$-th bit indicates the presence of the $i$-th artifact type, where the hard artifact is noted externally. For the clean reference images, this vector is strictly zero-valued ($[0,0,0,0,0]$).

\paragraph{Synthetic Detection Prompt}
For the synthetic ablation study, we employ a targeted prompt to verify the presence of specific artifacts. We instruct the model to act as an "expert biometric image quality analyst" and analyze the image for a predefined list of issues: \textit{"blur, noise, pixelation (low\_res), jpeg artifacts (compression), and bad lighting (under/overexposed)."} To ensure the outputs are machine-readable, the prompt explicitly commands the model to \textit{"Return STRICT JSON"} containing a scalar \textit{quality\_score} (0--100) and boolean detection flags (e.g., \textit{has\_blur}, \textit{has\_noise}) for each degradation type.

\paragraph{Evaluation Metrics}
We evaluate the VLM's predicted degradation tags against these ground truth vectors using three specialized metrics:
\begin{enumerate}
    \item \textbf{Hallucination Rate}: Estimates the model's tendency to predict degradations on the clean reference images (False Positive Rate).
    \item \textbf{Degradation Recall}: Measures the model's ability to correctly identify the designated "Hard Artifact" within the mixed variant, identifying the primary cause of quality loss.
    \item \textbf{Hamming Precision}: Quantifies how accurately the model describes the complete degradation state by measuring the Hamming distance between the predicted and ground truth vectors.
\end{enumerate}

\section{Experimental Results}
\label{sec:experiment}

In this section, we present the zero-shot FIQA performance of diverse VLMs. We organize our evaluation into three parts: first, we benchmark biometric utility on real-world datasets; second, we analyze internal consistency and prompt sensitivity; and finally, we conduct a synthetic ablation to assess degradation-detection capabilities in a controlled environment. In addition to our systematic analysis, we provide a qualitative examination of sample VLM outputs to visually demonstrate their interpretability.

\subsection{Experimental Setup}
\paragraph{Models} 
We evaluate a broad selection of state-of-the-art open-source VLMs to assess the impact of architecture and scale on FIQA performance. We analyze the QWEN Family (QWEN2-7B, QWEN2.5-7B/32B/72B)~\cite{wang2024qwen2,bai2025qwen2} to observe scaling trends, Gemma-3-4B~\cite{team2025gemma}, Idefics-9B~\cite{laurencon2023obelics}, and Phi-4-6B~\cite{abdin2024phi} as other lightweight alternatives.

\paragraph{Baselines}
We compare these zero-shot estimators against learning-based FIQA methods: eDifFIQA~\cite{babnik2024ediffiqa}, FaceQAN~\cite{babnik2022faceqan}, SDD-FIQA~\cite{ou2021sddfiqa}, and ViT-FIQA~\cite{atzori2025vitfiqa}.

\subsection{Biometric Utility Analysis (Error-versus-Reject)}
\label{subsec:evr_analysis}

\begin{figure}[!t]
  \centering
  \begin{subfigure}{0.9\linewidth}
    \includegraphics[width=\linewidth]{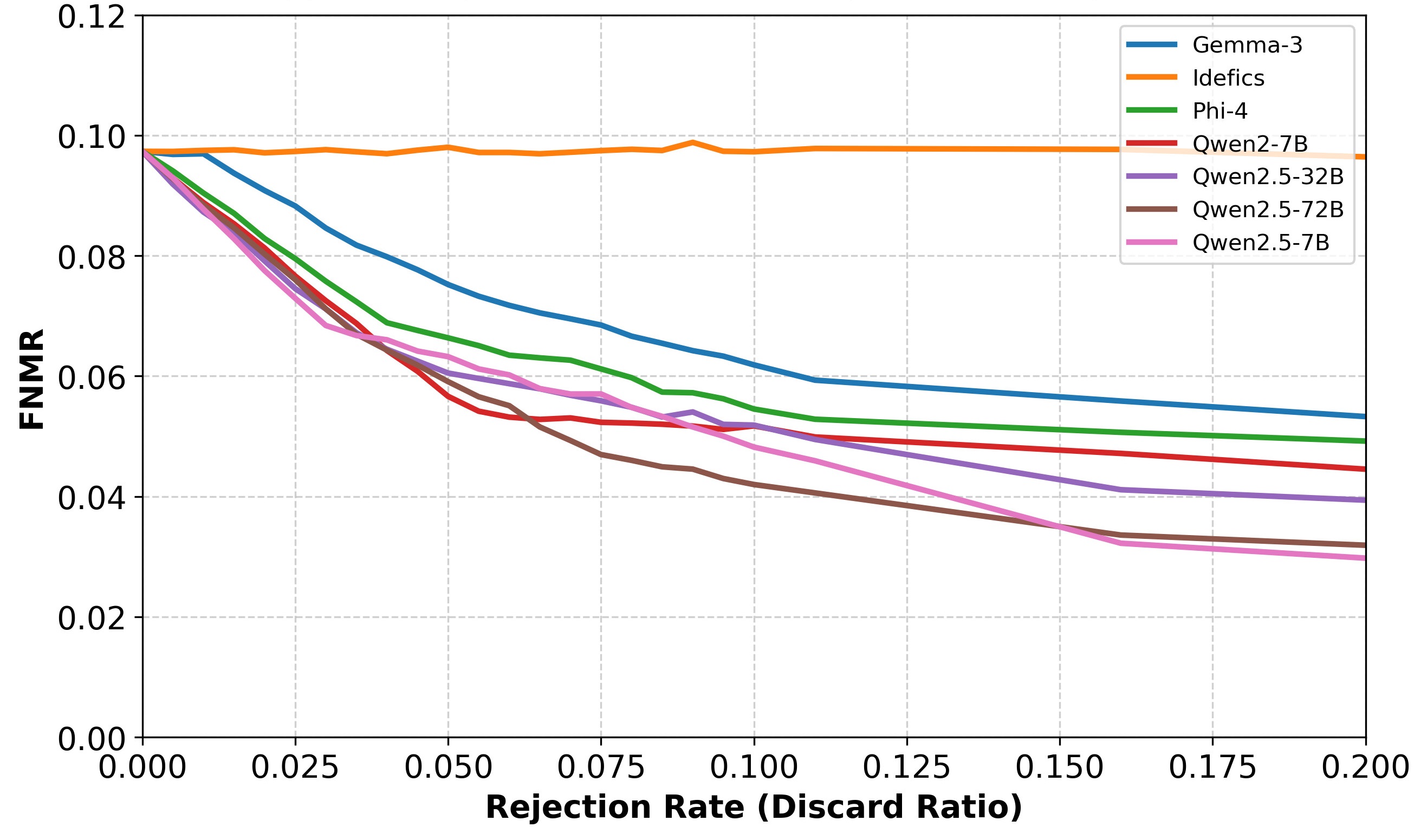}
    \caption{Comparison of different VLM Architectures (Simple Prompt)}
  \end{subfigure}
  \hfill
  \begin{subfigure}{0.9\linewidth}
    \includegraphics[width=\linewidth]{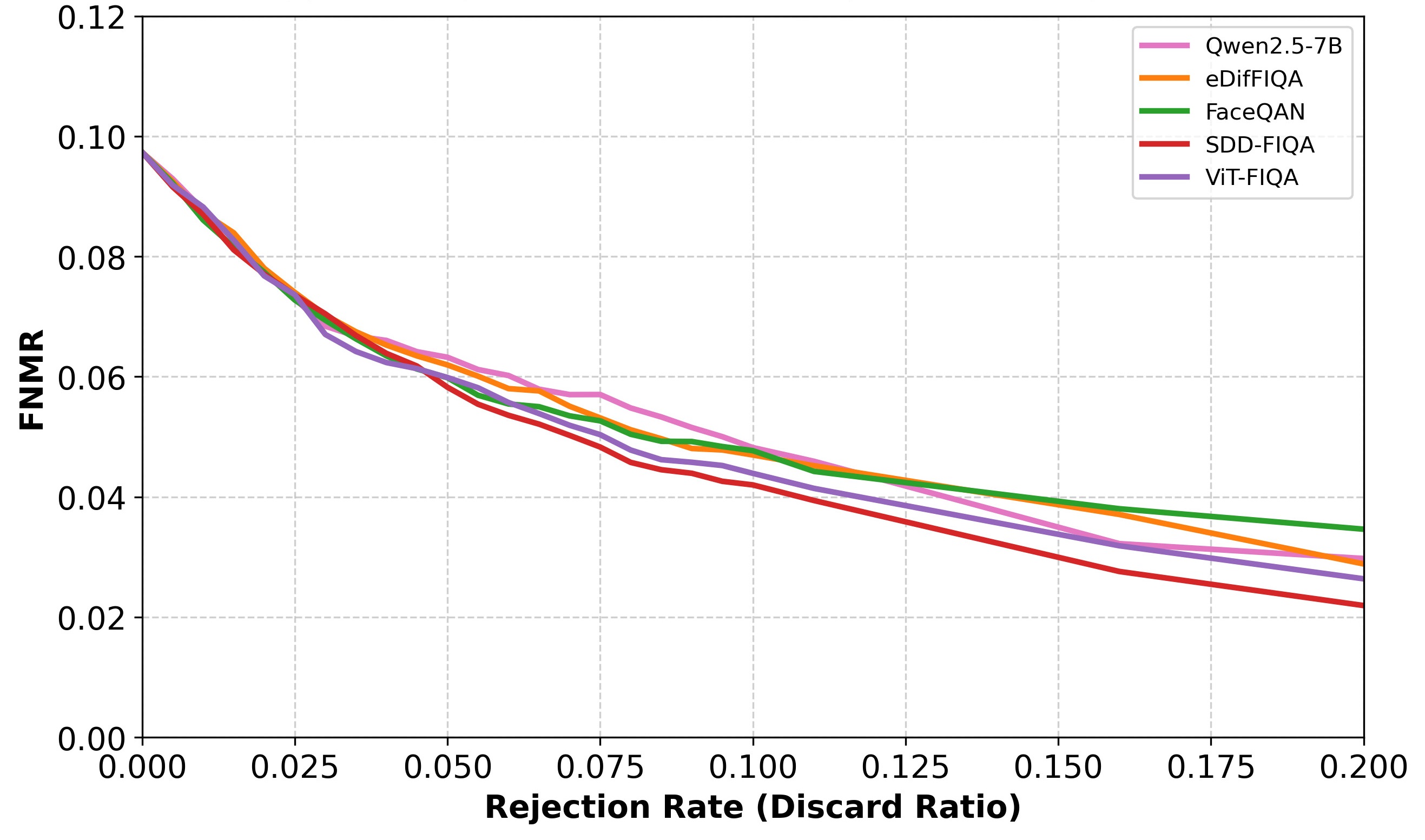}
    \caption{QWEN2.5-7B vs. SOTA Baselines}
  \end{subfigure}
  \caption{Error-versus-Reject (EvR) curves on LFW.}
  \label{fig:evr_curves}
\end{figure}

We first evaluate the primary requirement of FIQA: the ability to filter out samples that cause recognition errors. Fig.~\ref{fig:evr_curves} presents the error-versus-reject (EvR) curves on the LFW dataset (ArcFace~\cite{deng2019arcface} embeddings, FMR=$10^{-3}$).

Quantitative results are detailed in Table~\ref{tab:evr_main}, where AUC scores are averaged across multiple recognition backbones (ArcFace~\cite{deng2019arcface}, TransFace~\cite{dan2023transface}, and LVFace~\cite{you2025lvface}) to ensure the metrics are not biased toward a specific FR architecture.

\paragraph{VLM Performance}
Fig.~\ref{fig:evr_curves} demonstrates that most VLMs can replicate the behavior of FIQA methods in the zero-shot setting. Particularly, QWEN2.5-7B shows a curve that aligns closely with these methods. We can see similar results in Table~\ref{tab:evr_main}. Regarding AUC (@ 10\% and 20\%), QWEN2.5-72B not only outperforms other VLM variants but also surpasses the specialized baselines, except SDD-FIQA. This indicates that VLMs' advanced internal reasoning can effectively proxy biometric quality estimation without task-specific supervision.

\paragraph{Scaling and Architecture}
The QWEN family results in Table~\ref{tab:evr_main} reveal a complex relationship between scale and utility that varies by rejection depth. While the 72B model dominates in the broader range (best AUC @ 10\% and 20\%), the smaller 7B variant surprisingly proves more effective in the high-utility regime, achieving the best VLM performance at 5\%. Likewise, the 7B model consistently outperforms the mid-sized 32B variant across most metrics. Additionally, we observe a significant performance gap across architectures; despite comparable sizes, Gemma-3 and Idefics lag behind the QWEN series, underscoring that the underlying architectural choices are more critical than parameter count alone.

\begin{table}[!t]
\centering
\caption{Main Benchmark: Biometric Utility. We report the accumulated partial AUC (lower is better). VLMs are evaluated using the standard 'Simple' prompt.}
\label{tab:evr_main}
\footnotesize 
\setlength{\tabcolsep}{2pt} 
\renewcommand{\arraystretch}{1.1}
\begin{tabular}{|l||c|c|c|c|}
\hline
Method & AUC @ 1\% & AUC @ 5\% & AUC @ 10\% & AUC @ 20\% \\
\hline
\hline
\multicolumn{5}{|l|}{\textit{Supervised / Specialized Baselines}} \\
\hline
SDD-FIQA & 0.00091 & 0.00376 & 0.00622 & 0.00833 \\
ViT-FIQA & 0.00092 & 0.00372 & 0.00627 & 0.00856 \\
FaceQAN & 0.00092 & 0.00375 & 0.00638 & 0.00891 \\
eDifFIQA & 0.00092 & 0.00381 & 0.00649 & 0.00903 \\
\hline
\multicolumn{5}{|l|}{\textit{Vision-Language Models (Zero-Shot - Simple Prompt)}} \\
\hline
QWEN2.5-72B & 0.00092 & 0.00381 & 0.00624 & 0.00850 \\
QWEN2.5-32B & 0.00092 & 0.00379 & 0.00659 & 0.00936 \\
QWEN2.5-7B & 0.00092 & 0.00378 & 0.00657 & 0.00898 \\
QWEN2-7B & 0.00092 & 0.00382 & 0.00644 & 0.00937 \\
Phi-4 & 0.00094 & 0.00399 & 0.00703 & 0.01016 \\
Gemma-3 & 0.00097 & 0.00437 & 0.00779 & 0.01127 \\
Idefics & 0.00097 & 0.00485 & 0.00971 & 0.01555 \\
\hline
\end{tabular}
\end{table}

\subsection{Sensitivity to Surveillance Degradation}
\label{subsec:scface_analysis}

A robust quality estimator must reflect physical degradation. We use the SCFace dataset to measure how scores change across three standoff distances: Far ($d1$, 4.2~m), Medium ($d2$, 2.6~m), and Close ($d3$, 1.0~m).

\begin{figure}[!b]
  \centering
  \begin{subfigure}{0.85\linewidth}
    \includegraphics[width=\linewidth]{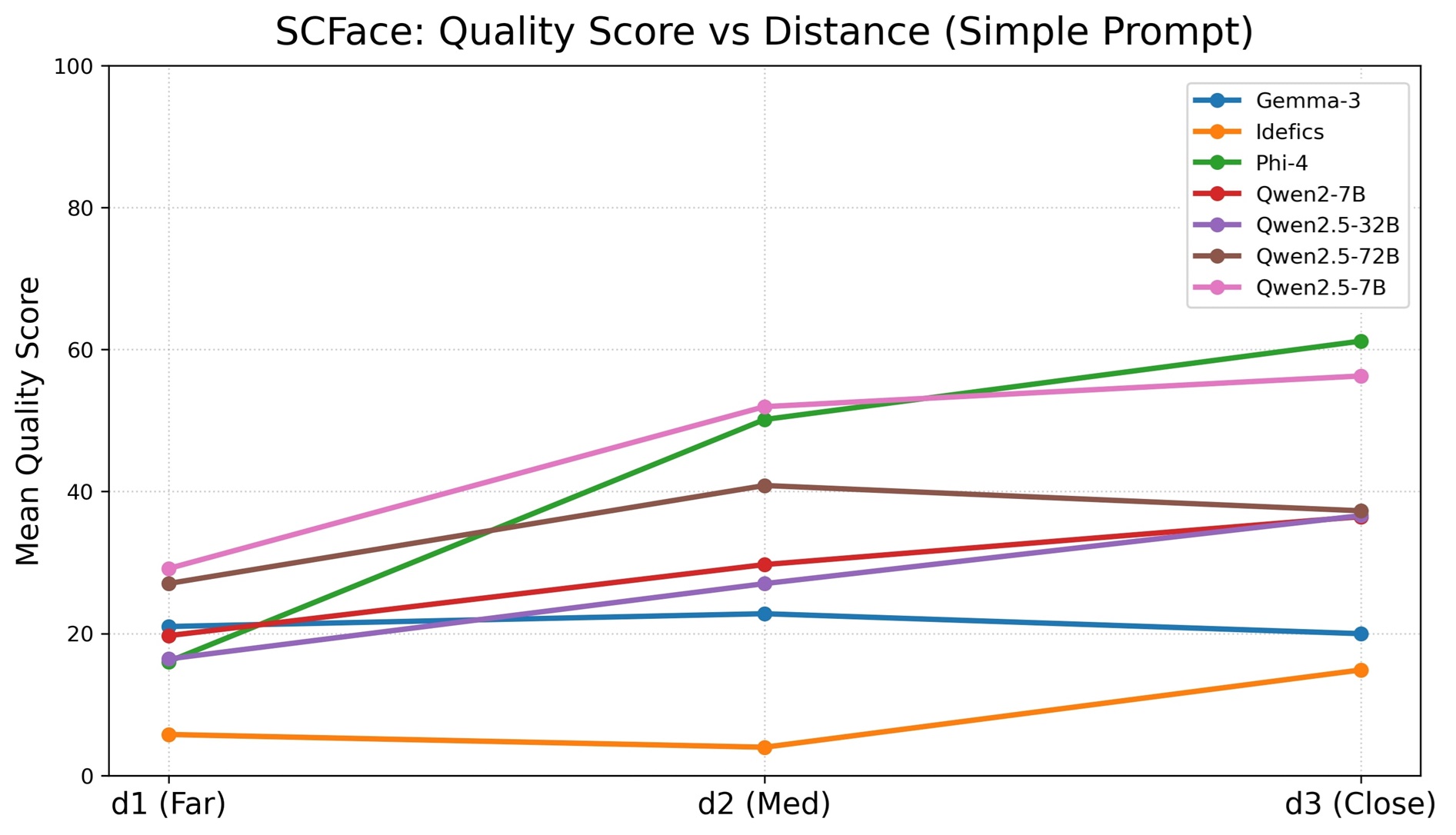}
    \caption{Global Model Comparison (Simple Prompt)}
  \end{subfigure}
  \hfill
  \begin{subfigure}{0.6\linewidth}
    \includegraphics[width=\linewidth]{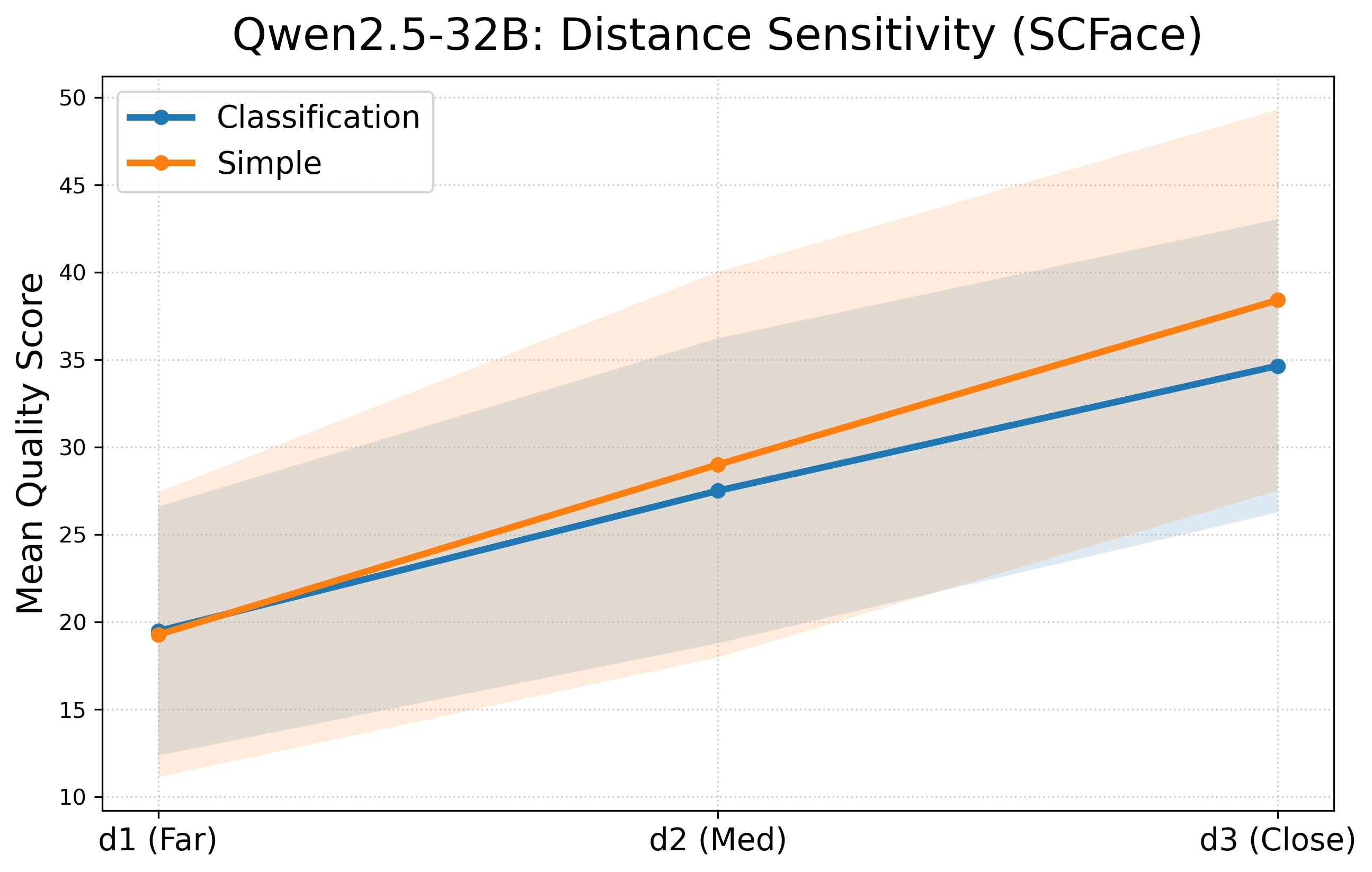}
    \caption{Detailed Trend for QWEN2.5-32B}
  \end{subfigure}
  \caption{Score sensitivity to physical distance in surveillance scenarios (SCFace). (a) Mean quality scores across architectures. (b) Score distributions for QWEN2.5-32B across the three distances.}
  \label{fig:scface_sensitivity}
\end{figure}

The impact of distance on model scoring is shown in Fig.~\ref{fig:scface_sensitivity}. While the QWEN family shows the expected monotonic increase in quality as the subject approaches the camera, other models struggle to differentiate the scenarios (Fig.~\ref{fig:scface_sensitivity}a). Specifically, QWEN2.5-32B demonstrates a sharp distinction between distances (Fig.~\ref{fig:scface_sensitivity}b), with mean quality scores rising consistently: $\approx16.5$ (Far), $\approx27.0$ (Med), $\approx36.5$ (Close). This confirms the model correctly perceives the loss of facial detail. In contrast, Gemma-3 produces nearly flat scores over distances (around $20.0$), whereas Idefics assigns 0 to almost all images at the Far and Med distances.

\begin{figure*}[!t]
  \centering
  \includegraphics[width=\linewidth]{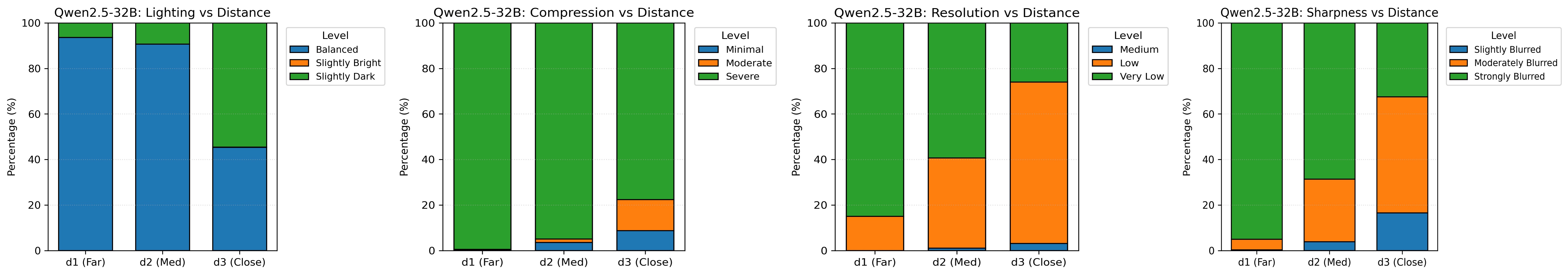}
  \caption{Explainability analysis on SCFace (QWEN2.5-32B). The plot shows the distribution of generated attribute labels across the three camera distances.}
  \label{fig:scface_attributes}
\end{figure*}

Understanding why VLMs reduce the importance of distant faces involves analyzing the attribute classification outputs of QWEN2.5-32B, as illustrated in Fig.~\ref{fig:scface_attributes}. For $d1$ (Far) images, the model assigns higher probabilities to "Low Resolution" and "Blurry." As the subject moves closer, the distribution of generated attribute labels changes consistently, confirming that lower scores are driven by recognized physical degradation rather than arbitrary noise.

\subsection{Prompt Consistency and Robustness}
\label{subsec:consistency}

We examine whether changing the prompt phrasing alters the model's ranking logic. We compare the base "Simple" prompt against "Utility," "Reliability," and "Classification" variants across different model scales.

\begin{figure}[!b]
  \centering
  \includegraphics[width=0.45\linewidth]{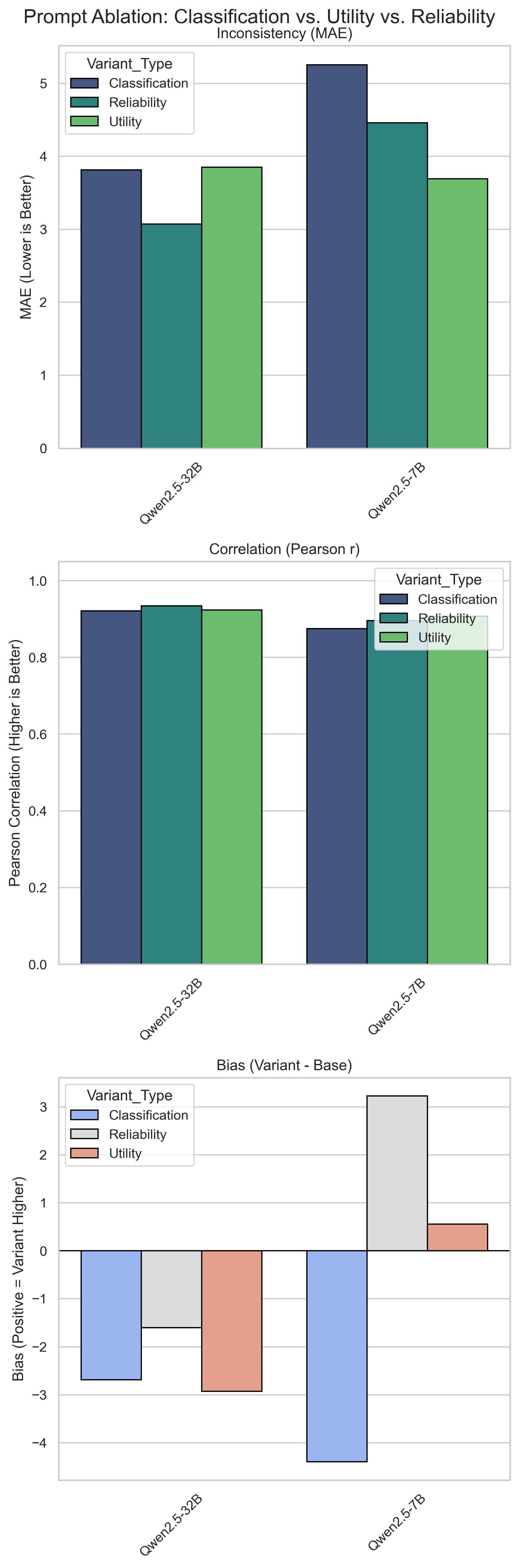}
  \caption{Prompt Ablation Study: Comparison of score distributions between the Simple Quality prompt and Classification Quality, Utility, and Reliability prompt variants.}
  \label{fig:prompt_ablation}
\end{figure}

Fig.~\ref{fig:prompt_ablation} suggests that semantic phrasing introduces minor calibration offsets while preserving the underlying quality scale. The "Classification" prompt tends to result in slightly stricter scoring (negative bias) across architectures. However, model scale affects optimal phrasing: the larger 32B model is most consistent with "Reliability," whereas the smaller 7B model aligns best with "Utility."

Table~\ref{tab:evr_prompts} presents the AUC performance for prompt variants. For the larger QWEN2.5-32B, the internal ranking logic appears highly robust, showing a possible invariance to semantic signals. In contrast, the smaller QWEN2.5-7B is more affected by prompting. Both "Reliability" and "Utility" variants outperform the base prompt, suggesting that smaller architectures may require explicit semantic cues to effectively activate their biometric-aligned features.

The "Classification" strategy yields divergent results across model families and scales. While it improves the QWEN2-7B and remains effective for the QWEN2.5-32B model, it degrades the performance of newer small-scale models like QWEN2.5-7B and Phi-4. This indicates that while asking for categorical labels can stabilize certain architectures, lightweight models may lose fine-grained ranking resolution when forced into a discrete classification mode.

\begin{table}[!b]
\centering
\caption{Effect of Prompt Phrasing on Biometric Utility (accumulated partial AUC). Comparing all prompt strategies. Lower is better.}
\label{tab:evr_prompts}
\footnotesize 
\setlength{\tabcolsep}{2pt} 
\renewcommand{\arraystretch}{1.1}

\begin{tabular}{|l|l||c|c|c|c|}
\hline
\textbf{Model} & \textbf{Variant} & \textbf{AUC} & \textbf{AUC} & \textbf{AUC} & \textbf{AUC} \\
 & & \textbf{@ 1\%} & \textbf{@ 5\%} & \textbf{@ 10\%} & \textbf{@ 20\%} \\
\hline
\hline
\multirow{2}{*}{QWEN2-7B} 
 & Simple & 0.00092 & 0.00382 & 0.00644 & 0.00937 \\
 & Classif. & 0.00093 & 0.00377 & 0.00614 & 0.00816 \\
\hline
\multirow{4}{*}{QWEN2.5-7B} 
 & Simple & 0.00092 & 0.00378 & 0.00657 & 0.00898 \\
 & Reliability & 0.00092 & 0.00378 & 0.00628 & 0.00849 \\
 & Utility & 0.00092 & 0.00375 & 0.00625 & 0.00845 \\
 & Classif. & 0.00092 & 0.00376 & 0.00658 & 0.00960 \\
\hline
\multirow{4}{*}{QWEN2.5-32B} 
 & Simple & 0.00092 & 0.00379 & 0.00659 & 0.00936 \\
 & Reliability & 0.00092 & 0.00379 & 0.00664 & 0.00961 \\
 & Utility & 0.00092 & 0.00382 & 0.00652 & 0.00933 \\
 & Classif. & 0.00092 & 0.00376 & 0.00653 & 0.00921 \\
\hline
\multirow{2}{*}{Phi-4} 
 & Simple & 0.00094 & 0.00399 & 0.00703 & 0.01016 \\
 & Classif. & 0.00094 & 0.00425 & 0.00741 & 0.01069 \\
\hline
\end{tabular}
\end{table}

\subsection{Internal Consistency}
\label{subsec:internal_consistency}

\begin{figure*}[!t]
  \centering
  \includegraphics[width=\linewidth]{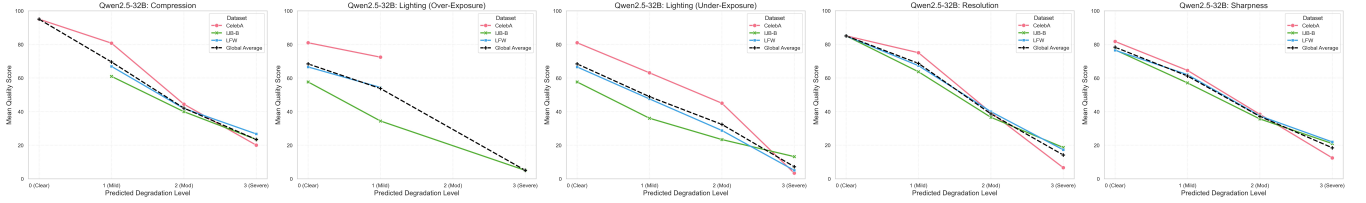}
  \caption{Internal Consistency (QWEN2.5-32B). Boxplots of scalar quality scores grouped by the model's generated text labels.}
  \label{fig:internal_consistency}
\end{figure*}

Finally, we assess the "Internal Consistency" of the VLM. A valid explainer must be consistent with its own scores: if the model labels some images as "Blurry" or "Low Resolution," the average score assigned to those images must be lower than that of images labeled "Clear".

Fig.~\ref{fig:internal_consistency} validates this behavior for QWEN2.5-32B, demonstrating a strictly monotonic connection between generated text labels and scalar scores. By plotting the mean quality score against the model's predicted degradation levels across datasets (CelebA, IJB-B, LFW) and their global average, we observe a trend: as the textual description shifts from "Clear" to "Severe," the scalar utility drops greatly. This alignment is universal across all evaluated dimensions—Compression, Lighting, Resolution, and Sharpness—where images labeled "Clear" consistently maintain high global averages ($\approx 80$--$95$), while those flagged as "Severe" fall to the $5$--$25$ range. This confirms that the scalar score is not an arbitrary hallucination but is semantically grounded in the model's explicit perception of visual artifacts.

\subsection{Synthetic Ablation: Mix Degradation}
\label{subsec:synthetic_ablation}

To further dissect the model's behavior, we evaluate its performance on our \textbf{Mix Degradation Benchmark}. This test measures three critical dimensions: trustworthiness on clean images ($L_0$), the ability to detect severe degradations in complex mixtures ($L_2$), and the precision of categorical outputs (Hamming Distance).

\paragraph{Clean Image Trustworthiness ($L_0$)}
We first evaluate the model's tendency to predict degradations on perfectly clean images (False Positive rate). As shown in Table~\ref{tab:mix_l0}, QWEN2.5-7B emerges as the most reliable detector, correctly identifying 87.9\% of clean images as degradation-free. However, its scoring logic is erratic; when it hallucinates a single degradation, the quality score collapses disproportionately and behaves non-monotonically as errors increase. In contrast, while the larger QWEN2.5-32B notices artifacts (hallucinating in $\approx 50\%$ of samples), it maintains superior internal consistency; the quality scores decrease consistently as more false features appear. 

\begin{table}[!b]
\centering
\caption{L0 Analysis (Clean Images): False Positive (FP) rates and associated Quality Scores (QS).}
\label{tab:mix_l0}
\footnotesize
\setlength{\tabcolsep}{2pt} 
\renewcommand{\arraystretch}{1.1}
\begin{tabular}{|l||c|c||c|c||c|c||c|c|}
\hline
\multirow{2}{*}{\textbf{Model}} & \multicolumn{2}{c||}{\textbf{0 FP}} & \multicolumn{2}{c||}{\textbf{1 FP}} & \multicolumn{2}{c||}{\textbf{2 FP}} & \multicolumn{2}{c|}{\textbf{$\ge$3 FP}} \\
\cline{2-9}
 & \textbf{\%} & \textbf{QS} & \textbf{\%} & \textbf{QS} & \textbf{\%} & \textbf{QS} & \textbf{\%} & \textbf{QS} \\
\hline
\hline
QWEN2.5-7B & 87.9 & 94.8 & 11.1 & 33.6 & 0.9 & 44.3 & 0.2 & 40.3 \\
QWEN2-7B & 75.7 & 90.1 & 18.8 & 43.8 & 3.0 & 46.3 & 2.5 & 29.7 \\
QWEN2.5-32B & 50.9 & 95.0 & 7.5 & 82.9 & 39.2 & 69.6 & 2.4 & 44.6 \\
\hline
\end{tabular}
\end{table}

\paragraph{Degradation Detection \& Recall ($L_2$)}
In the mixed degradation scenario, we evaluate the model's ability to isolate a "Hard" artifact from milder degradations. Table~\ref{tab:mix_l2_combined} summarizes the detection completeness and "Hard" artifact recall. QWEN2-7B proves to be the most sensitive artifact detector, achieving the highest recall for the "Hard" artifact (79.8\%) and the best completeness (identifying all 3 degradations in 39.7\% of cases). QWEN2.5-7B follows as a balanced alternative in general. Conversely, the larger QWEN2.5-32B fails significantly in this dense detection task; it successfully identifies all three degradations in only 8.7\% of cases, often overlooking the severe artifact entirely.

\begin{table}[!b]
\centering
\caption{L2 Analysis (Mixed Degradation): Detection Completeness and Hard Artifact Recall.}
\label{tab:mix_l2_combined}
\footnotesize
\setlength{\tabcolsep}{2pt} 
\renewcommand{\arraystretch}{1.1}

\begin{tabular}{|l||c|c|c|c||c|}
\hline
\multirow{2}{*}{\textbf{Model}} & \multicolumn{4}{c||}{\textbf{Completeness}} & \textbf{Hard Artifact} \\
\cline{2-5} 
 & \textbf{0/3} & \textbf{1/3} & \textbf{2/3} & \textbf{3/3} & \textbf{Recall (\%)} \\
\hline
\hline
QWEN2-7B & 2.2 & 12.2 & 45.9 & 39.7 & 79.8 \\
QWEN2.5-7B & 3.6 & 32.3 & 27.4 & 36.7 & 71.3 \\
QWEN2.5-32B & 1.7 & 41.8 & 47.8 & 8.7 & 62.0 \\
\hline
\end{tabular}
\end{table}

\paragraph{Output Precision (Hamming Distance)}
Finally, we evaluate the exactness of the predicted degradation vectors by analyzing the Hamming prediction errors ($D_H$) in Table~\ref{tab:mix_hamming}. QWEN2.5-32B achieves the best low-error performance, securing the highest rates for perfect matches ($D_0=7.8\%$) and single-bit errors ($D_1=32.1\%$). However, its error distribution is broad, with significant mass spilling into high-error categories ($D_3=23.2\%$). In contrast, the 7B models exhibit a rigid, systematic bias: they peak sharply at $D_2$ ($\approx 55\%$) while keeping severe errors ($D_3$) relatively low. This suggests that smaller models deterministically miss specific artifact pairs (likely mild ones), whereas the larger model is more capable but less predictable.

\begin{table}[!b]
\centering
\caption{Hamming Distance Breakdown: Distribution of prediction errors ($D_H$).}
\label{tab:mix_hamming}
\footnotesize
\setlength{\tabcolsep}{2pt} 
\renewcommand{\arraystretch}{1.1}

\begin{tabular}{|l||c|c|c|c|c|c|}
\hline
\textbf{Model} & \textbf{D0 (Perf.)} & \textbf{D1} & \textbf{D2} & \textbf{D3} & \textbf{D4} & \textbf{D5} \\
\hline
\hline
QWEN2.5-32B & 7.8 & 32.1 & 31.8 & 23.2 & 3.7 & 1.5 \\
QWEN2-7B & 3.5 & 21.2 & 58.6 & 14.4 & 2.3 & 0.0 \\
QWEN2.5-7B & 3.1 & 26.2 & 54.1 & 15.2 & 1.3 & 0.0 \\
\hline
\end{tabular}
\end{table}

\subsection{Sample Outputs}
\label{subsec:sample_outputs}

\begin{figure*}[!h]
    \centering
    \begin{subfigure}[b]{0.425\textwidth}
        \centering
        \includegraphics[width=\textwidth]{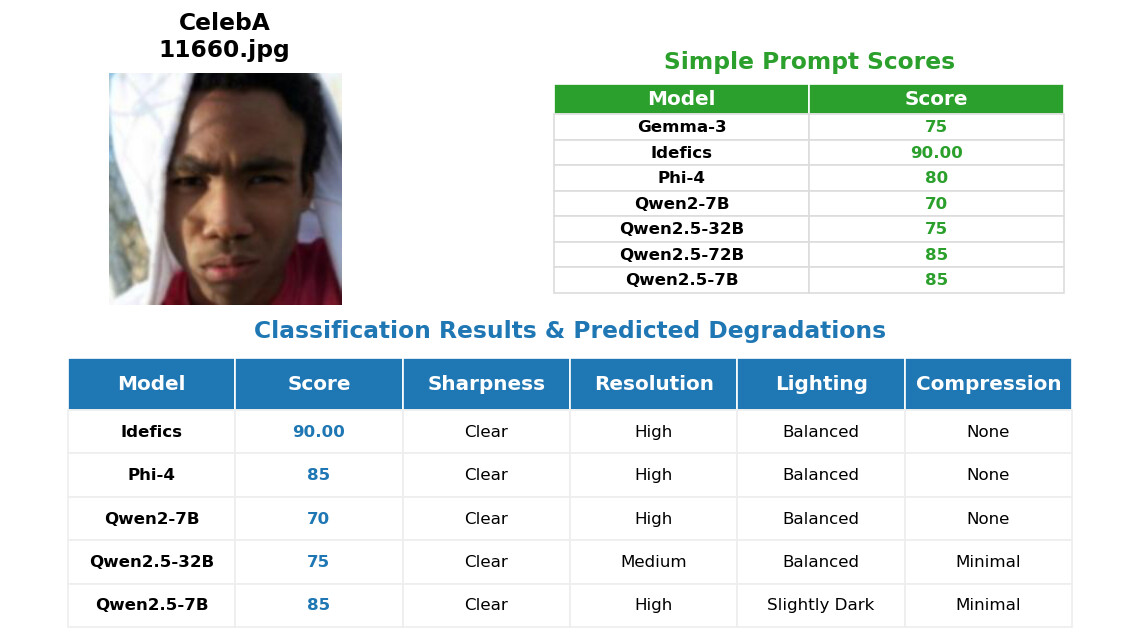}
        \caption{CelebA-HQ}
        \label{fig:top-left}
    \end{subfigure}
    \hfill 
    \begin{subfigure}[b]{0.425\textwidth}
        \centering
        \includegraphics[width=\textwidth]{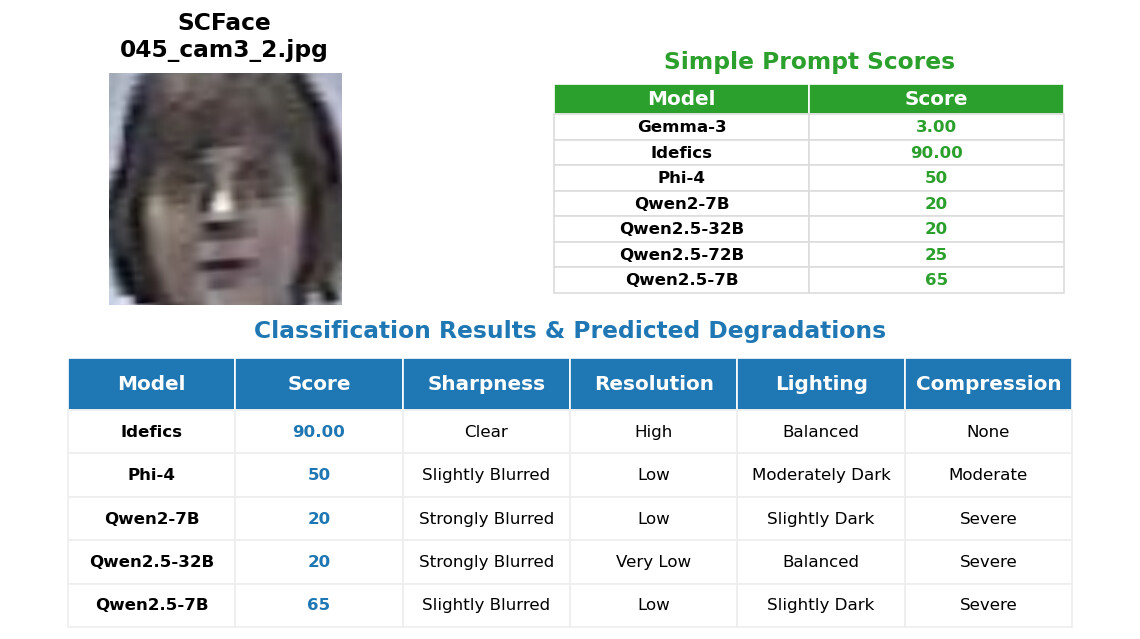}
        \caption{SCFace}
        \label{fig:top-right}
    \end{subfigure}
    
    
    \vspace{0.5cm} 
    
    \begin{subfigure}[b]{0.425\textwidth}
        \centering
        \includegraphics[width=\textwidth]{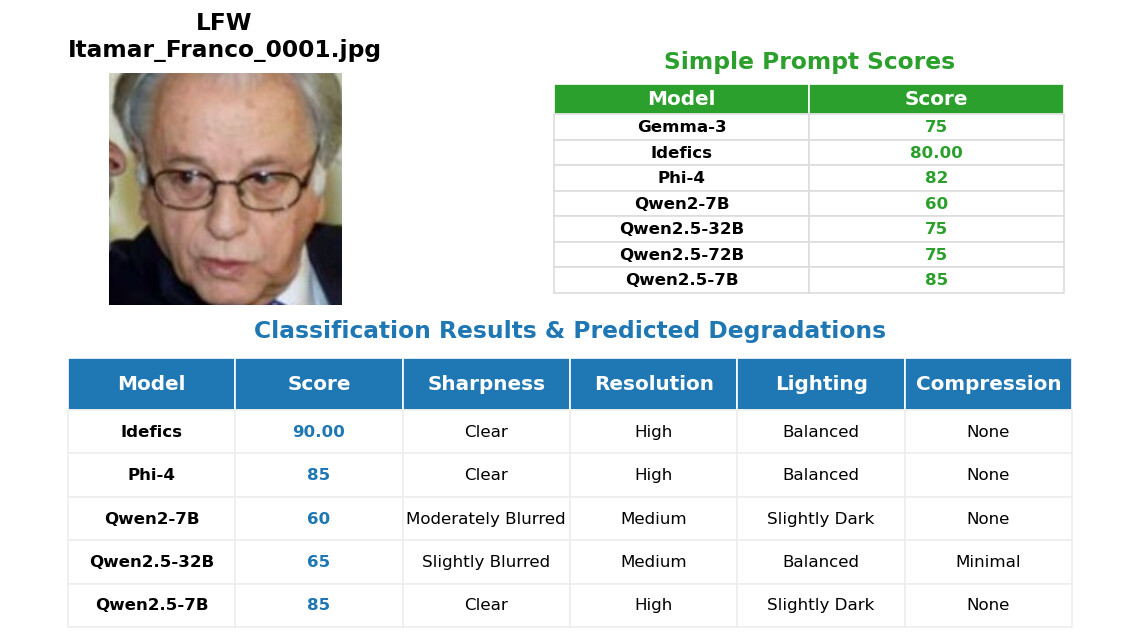}
        \caption{LFW}
        \label{fig:bottom-left}
    \end{subfigure}
    \hfill
    \begin{subfigure}[b]{0.425\textwidth}
        \centering
        \includegraphics[width=\textwidth]{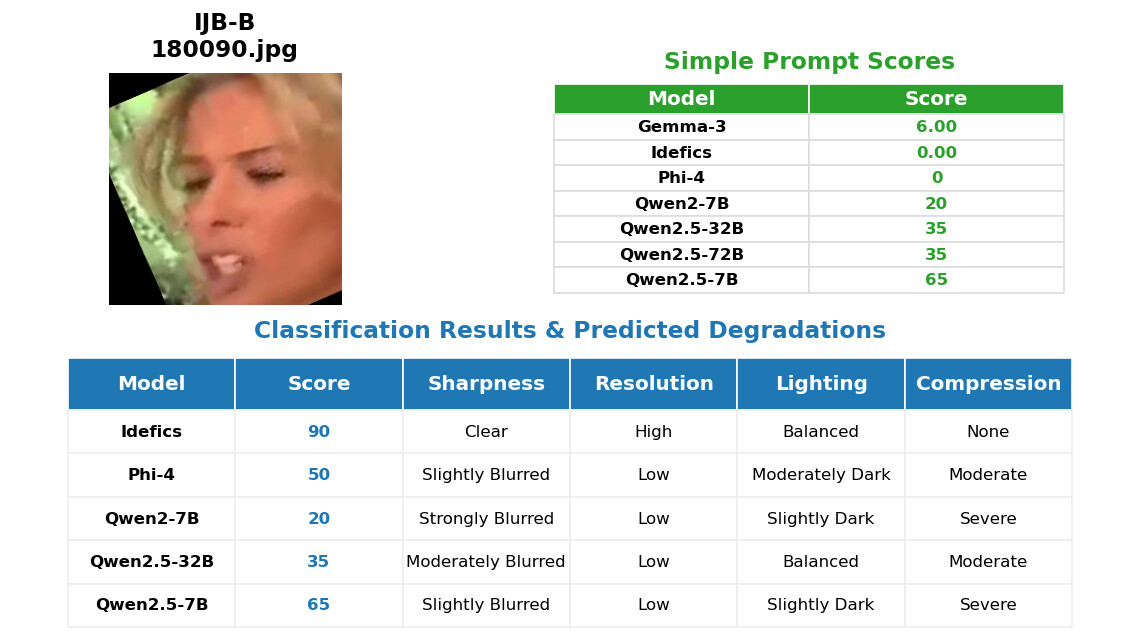}
        \caption{IJB-B}
        \label{fig:bottom-right}
    \end{subfigure}
    
    \caption{Sample output of VLMs from all four datasets.}
    \label{fig:four-images}
\end{figure*}

We visualize sample VLM outputs across the four benchmark datasets in Fig.~\ref{fig:four-images} to provide a qualitative perspective.
The figure highlights the model's ability to generate both scalar quality scores and semantic attribute classifications. 
For the high-quality reference sample from CelebA-HQ (Fig.~\ref{fig:four-images}a), the models consistently assign high quality scores and correctly identify the attributes as "Clear" and "High Resolution." 
In contrast, for the surveillance sample from SCFace (Fig.~\ref{fig:four-images}b), the models successfully detect severe degradations, flagging the image as blurry or low resolution and attenuating the quality score accordingly. This qualitative evidence affirms our quantitative findings, demonstrating that VLMs can effectively translate visual artifacts into interpretable textual descriptions.

\section{Limitations}
\label{sec:limitations}

While our findings demonstrate the strong zero-shot capabilities of VLMs for FIQA, a primary limitation is their computational latency. Our latency analysis using a batch size of 16 on an RTX 4090 GPU shows that even a fast VLM (QWEN2.5-7B) with a simple prompt takes 33.3 ms/image, which is nearly 50$\times$ slower than a classical baseline like eDifFIQA (0.7 ms/image). Switching to different VLM architectures makes the process approximately 3 times slower; for instance, Gemma-3 and Phi-4 record latencies of 104.9 ms/image and 111.4 ms/image, respectively. The same 3-fold slowdown holds true for generating more descriptive text: using a longer, classification-based prompt on QWEN2.5-7B increases latency to 102.7 ms/image. Consequently, VLMs are too slow for high-throughput pipelines and should serve solely as a supplementary mechanism. Beyond computational costs, VLMs also struggle to reliably differentiate between high- and very high-quality images. Finally, our synthetic ablation reveals that larger models exhibit a hallucination rate of up to 50\% on clean samples, indicating that model size is not directly correlated with the FIQA performance.

\section{Conclusion}
\label{sec:conclusion}

In this work, we investigated the potential of off-the-shelf Vision-Language Models (VLMs) to serve as zero-shot estimators for Face Image Quality Assessment (FIQA). To address the interpretability limitations of traditional "black-box" methods, we introduced a comprehensive evaluation framework prioritizing biometric utility. We benchmarked a diverse set of open-source models against traditional FIQA baselines using error-versus-reject (EvR) curves. Furthermore, we examined the models' interpretability and consistency through qualitative analysis, including synthetic ablation and surveillance-specific scenarios. We also provided example VLM outputs to demonstrate their behavior in sample face images. Based on our comprehensive analysis, we derived the following key observations regarding the viability of VLMs for FIQA:

\begin{itemize}
    \item \textbf{Observation 1: Biometric utility depends on architecture, not just parameter count.} Our evaluation indicates that a VLM's ability to proxy biometric utility depends significantly on architectural choices, rather than purely on parameter count. We observed that, while specific model families, such as QWEN, achieved competitive alignment with specialized FIQA methods, other models with similar capacities missed key quality cues. This suggests that zero-shot FIQA capability is not universal across all large-scale models.
    
    \item \textbf{Observation 2: VLMs effectively bridge the interpretability gap.} Regarding interpretability, our analysis reveals that capable VLMs can provide the descriptive feedback often missing in current pipelines. In surveillance scenarios, these models not only downgraded low-quality samples but also provided textual justifications, such as "Low Resolution" or "Blur", that accurately reflected the visual degradation. This transparency underscores the potential of VLMs to support human-in-the-loop applications that require actionable feedback.
    
    \item \textbf{Observation 3: There is a complex trade-off between descriptive precision and scoring consistency.} Our synthetic ablation revealed a complex trade-off between scoring utility and descriptive precision. We found that larger models frequently hallucinate degradations on clean images. Yet, unlike their smaller counterparts, they maintained strict internal consistency, with scalar scores degrading monotonically with the severity of the generated critique. This implies that while larger models may be hyper-critical, their scoring logic appears more rational and predictable than the erratic behavior observed in smaller architectures.
    
    \item \textbf{Observation 4: Robustness to prompt phrasing scales inversely with model size.} We observed that prompt sensitivity tends to scale inversely with model size. While the ranking logic of larger models was robust to phrasing changes, smaller architectures were more strongly affected by targeted "Utility" or "Reliability" prompts. This indicates that smaller models likely benefit from specific semantic triggers to align their features with biometric requirements, whereas larger models exhibit a more generalized and stable concept of quality.
\end{itemize}

In future work, we plan to broaden our research by incorporating additional VLMs and evaluating their performance across a wider range of datasets. Additionally, we intend to integrate more biometric factors, such as expression, pose, and occlusion, into the assessment pipeline. Finally, we aim to compile these findings into a comprehensive benchmarking scheme.

{\small
\bibliographystyle{ieee}
\bibliography{egbib}

@String(ECCV= {Eur. Conf. Comput. Vis.})

@String(ICPR = {Int. Conf. Pattern Recog.})

@String(AAAI = {AAAI})

@String(ECCV  = {ECCV})

@String(ICPR  = {ICPR})

@article{schlett2022face,
  title={Face Image Quality Assessment: A Literature Survey},
  author={Schlett, Thomas and Rathgeb, Christian and Henniger, Olaf and Galbally, Javier and Fierrez, Julian and Busch, Christoph},
  journal={ACM Computing Surveys},
  year={2022},
  volume={54},
  number={10s},
  pages={1--49}
}

@inproceedings{najafzadeh2023face,
  title={Face Image Quality Vector Assessment for Biometrics Applications},
  author={Najafzadeh, Nima and Kashiani, Hossein and Saadabadi, Mohammad Saeed Ebrahimi and Talemi, Niloufar Alipour and Malakshan, Sahar Rahimi and Nasrabadi, Nasser M},
  booktitle={Proceedings of the IEEE/CVF Winter Conference on Applications of Computer Vision},
  pages={511--520},
  year={2023}
}

@inproceedings{wang2023exploring,
  title={Exploring {CLIP} for Assessing the Look and Feel of Images},
  author={Wang, Jianyi and Chan, Kelvin CK and Loy, Chen Change},
  booktitle={Proceedings of the AAAI conference on Artificial Intelligence},
  volume={37},
  number={2},
  pages={2555--2563},
  year={2023}
}

@inproceedings{zhang2023blind,
  title={Blind Image Quality Assessment via Vision-Language Correspondence: A Multitask Learning Perspective},
  author={Zhang, Weixia and Zhai, Guangtao and Wei, Ying and Yang, Xiaokang and Ma, Kede},
  booktitle={Proceedings of the IEEE/CVF Conference on Computer Vision and Pattern Recognition},
  pages={14071--14081},
  year={2023}
}

@inproceedings{terhorst2020ser,
  title={{SER-FIQ}: Unsupervised Estimation of Face Image Quality Based on Stochastic Embedding Robustness},
    author={Terhorst, Philipp and Kolf, Jan Niklas and Damer, Naser and Kirchbuchner, Florian and Kuijper, Arjan},
  booktitle={Proceedings of the IEEE/CVF Conference on Computer Vision and Pattern Recognition},
  pages={5651--5660},
  year={2020}
}

@inproceedings{kim2022adaface,
  title={{AdaFace}: Quality Adaptive Margin for Face Recognition},
  author={Kim, Minchul and Jain, Anil K and Liu, Xiaoming},
  booktitle={Proceedings of the IEEE/CVF Conference on Computer Vision and Pattern Recognition},
  pages={18750--18759},
  year={2022}
}

@article{miyata2024zen,
  title={{ZEN-IQA}: Zero-Shot Explainable and No-Reference Image Quality Assessment with Vision Language Model},
  author={Miyata, Takahiro},
  journal={IEEE Access},
  year={2024},
  volume={12},
  pages={70973--70983}
}

@inproceedings{otroshi2025facellm,
  title={{FaceLLM}: A Multimodal Large Language Model for Face Understanding},
  author={Shahreza, Hatef Otroshi and Marcel, S{\'e}bastien},
  booktitle={Proceedings of the IEEE/CVF International Conference on Computer Vision},
  pages={3677--3687},
  year={2025}
}

@inproceedings{chaubey2025facellava,
  title={{Face-LLaVA}: Facial Expression and Attribute Understanding through Instruction Tuning},
  author={Chaubey, Ashutosh and Guan, Xulang and Soleymani, Mohammad},
  booktitle={Proceedings of the IEEE/CVF Winter Conference on Applications of Computer Vision},
  pages={2648--2660},
  year={2026}
}

@inproceedings{wang2025faceshield,
  title={{FaceShield}: Explainable Face Anti-Spoofing with Multimodal Large Language Models},
  author={Wang, Hongyang and Shi, Yichen and Tao, Zhuofu and Gao, Yuhao and Zhang, Liepiao and Lin, Xun and Feng, Jun and Yuan, Xiaochen and Yu, Zitong and Cao, Xiaochun},
  booktitle={Proceedings of the AAAI Conference on Artificial Intelligence},
  volume={40},
  number={12},
  pages={9811--9819},
  year={2026}
}

@article{zhang2024herm,
  title={{HERM}: Benchmarking and Enhancing Multimodal {LLMs} for Human-Centric Understanding},
  author={Li, Kai and Yang, Zheng and Zhao, Jie and Shen, Haoliang and Hou, Rui and Chang, Hong and Yu, Yadong and Chen, Xilin},
  journal={arXiv preprint arXiv:2410.06777},
  year={2024}
}

@inproceedings{karras2018progressive,
  title={Progressive Growing of {GANs} for Improved Quality, Stability, and Variation},
  author={Karras, Tero and Aila, Timo and Laine, Samuli and Lehtinen, Jaakko},
  booktitle={International Conference on Learning Representations},
  year={2018}
}

@inproceedings{huang2007labeled,
  title={Labeled Faces in the Wild: A Database for Studying Face Recognition in Unconstrained Environments},
  author={Huang, Gary B. and Mattar, Marwan and Berg, Tamara and Learned-Miller, Erik},
  booktitle={Workshop on Faces in 'Real-Life' Images: Detection, Alignment, and Recognition (ECCV Workshop)},
  year={2008}
}

@inproceedings{whitelam2017ijbb,
  title={{IARPA Janus Benchmark-B} Face Dataset},
  author={Whitelam, Cameron and Taborsky, Emma and Blanton, Austin and Maze, Brianna and Adams, Jocelyn and Miller, Tim and Kalka, Nathan and Jain, Anil K and Duncan, James A and Allen, Kristen and others},
  booktitle={Proceedings of the IEEE Conference on Computer Vision and Pattern Recognition Workshops},
  pages={90--98},
  year={2017}
}

@article{grgic2011scface,
  title={{SCface} -- Surveillance Cameras Face Database},
  author={Grgic, Mislav and Delac, Kresimir and Grgic, Sonja},
  journal={Multimedia tools and applications},
  volume={51},
  number={3},
  pages={863--879},
  year={2011},
  publisher={Springer}
}

@article{babnik2024ediffiqa,
  title={{eDifFIQA}: Towards Efficient Face Image Quality Assessment Based On Denoising Diffusion Probabilistic Models},
  author={Babnik, \v{Z}iga and Peer, Peter and \v{S}truc, Vitomir},
  journal={IEEE Transactions on Biometrics, Behavior, and Identity Science},
  year={2024},
  volume={6},
  number={4},
  pages={458--474}
}

@article{bai2025qwen2,
  title={{Qwen2.5-VL} Technical Report},
  author={Bai, Shuai and Chen, Kun and Liu, Xiang and Wang, Jun and Ge, Wei and Song, Shiji and Xu, Ying and Lin, Jintao},
  journal={arXiv preprint arXiv:2502.13923},
  year={2025}
}

@inproceedings{deng2019arcface,
  title={{ArcFace}: Additive Angular Margin Loss for Deep Face Recognition},
  author={Deng, Jiankang and Guo, Jia and Xue, Niannan and Zafeiriou, Stefanos},
  booktitle={Proceedings of the IEEE/CVF Conference on Computer Vision and Pattern Recognition},
  pages={4690--4699},
  year={2019}
}

@article{you2025edqa,
  title={Enhancing Descriptive Image Quality Assessment with a Large-Scale Multi-Modal Dataset},
  author={You, Zhiyuan and Gu, Jinjin and Cai, Xin and Li, Zheyuan and Zhu, Kaiwen and Dong, Chao and Xue, Tianfan},
  journal={IEEE Transactions on Image Processing},
  volume={34},
  pages={8201--8215},
  year={2025},
  publisher={IEEE}
}

@inproceedings{wu2025fvq,
  title={{FVQ-20K}: A Large-Scale Dataset and an {LMM}-based Method for Face Video Quality Assessment},
  author={Wu, Sijing and Li, Yunhao and Xu, Ziwen and Gao, Yixuan and Duan, Huiyu and Sun, Wei and Zhai, Guangtao},
  booktitle={Proceedings of the 33rd ACM International Conference on Multimedia},
  pages={6928--6937},
  year={2025}
}

@inproceedings{ou2025mrfiqa,
  title={{MR-FIQA}: Face Image Quality Assessment with Multi-Reference Representations from Synthetic Data Generation},
  author={Ou, Fu-Zhao and Li, Chongyi and Wang, Shiqi and Kwong, Sam},
  booktitle={Proceedings of the IEEE/CVF International Conference on Computer Vision},
  pages={12915--12925},
  year={2025}
}

@inproceedings{ou2024clibfiqa,
  title={{CLIB-FIQA}: Face Image Quality Assessment with Confidence Calibration},
  author={Ou, Fu-Zhao and Li, Chongyi and Wang, Shiqi and Kwong, Sam},
  booktitle={Proceedings of the IEEE/CVF Conference on Computer Vision and Pattern Recognition},
  pages={1694--1704},
  year={2024}
}

@inproceedings{wu2024qalign,
  title={{Q-Align}: Teaching {LMMs} for Visual Scoring via Discrete Text-Defined Levels},
  author={Wu, Haoning and Zhang, Zicheng and Zhang, Weixia and Chen, Chaofeng and Liao, Liang and Li, Chunyi and Gao, Yixuan and Wang, Annan and Zhang, Erli and Sun, Wenxiu and others},
  booktitle={Proceedings of the 41st International Conference on Machine Learning},
  pages={54015--54029},
  year={2024}
}

@inproceedings{kabbani2025faceoracle,
  title={{FaceOracle}: Chat with a Face Image Oracle},
  author={Kabbani, Wassim and Raja, Kiran and Ramachandra, Raghavendra and Busch, Christoph},
  booktitle={European Conference on Computer Vision Workshops},
  pages={210--226},
  year={2024},
  organization={Springer}
}

@inproceedings{gao2025mdfiqa,
  title={Multi-Dimensional Text-to-Face Image Quality Assessment Using {LLM}: Database and Method},
  author={Gao, Yixuan and Min, Xiongkuo and Han, Jinliang and Cao, Yuqin and Wu, Sijing and Dou, Yunze and Zhai, Guangtao},
  booktitle={Proceedings of the 33rd ACM International Conference on Multimedia},
  pages={6948--6957},
  year={2025}
}

@inproceedings{wu2024openended,
  title={Towards Open-ended Visual Quality Comparison}, 
  author={Wu, Haoning and Zhu, Hanwei and Zhang, Zicheng and Zhang, Erli and Chen, Chaofeng and Liao, Liang and Li, Chunyi and Wang, Annan and Sun, Wenxiu and Yan, Qiong and others},
  booktitle={European Conference on Computer Vision},
  pages={360--377},
  year={2024},
  organization={Springer}
}

@inproceedings{lin2025instructflip,
  title={{InstructFLIP}: Exploring Unified Vision-Language Model for Face Anti-spoofing},
  author={Lin, Kun-Hsiang and Tseng, Yu-Wen and Huang, Kang-Yang and Wu, Jhih-Ciang and Cheng, Wen-Huang},
  booktitle={Proceedings of the 33rd ACM International Conference on Multimedia},
  pages={2987--2996},
  year={2025}
}

@inproceedings{ma2025vquala,
  title={{VQualA} 2025 Challenge on Face Image Quality Assessment: Methods and Results},
  author={Ma, Sizhuo and Chen, Wei-Ting and Gao, Qiang and Wang, Jian and Zhou, Chris Wei and Sun, Wei and Zhang, Weixia and Cao, Linhan and Jia, Jun and Zhu, Xiangyang and others},
  booktitle={Proceedings of the IEEE/CVF International Conference on Computer Vision},
  pages={3448--3457},
  year={2025}
}

@inproceedings{radford2021learning,
  title={Learning Transferable Visual Models From Natural Language Supervision},
  author={Radford, Alec and Kim, Jong Wook and Hallacy, Chris and Ramesh, Aditya and Goh, Gabriel and Agarwal, Sandhini and Sastry, Girish and Askell, Amanda and Mishkin, Pamela and Clark, Jack and others},
  booktitle={International Conference on Machine Learning},
  pages={8748--8763},
  year={2021},
  organization={PmLR}
}

@article{zhang2016joint,
  title={Joint Face Detection and Alignment Using Multitask Cascaded Convolutional Networks},
  author={Zhang, Kaipeng and Zhang, Zhanpeng and Li, Zhifeng and Qiao, Yu},
  journal={IEEE Signal Processing Letters},
  volume={23},
  number={10},
  pages={1499--1503},
  year={2016},
  publisher={IEEE}
}

@article{wang2024qwen2,
  title={{Qwen2-VL}: Enhancing Vision-Language Model's Perception of the World at Any Resolution},
  author={Wang, Peng and Bai, Shuai and Tan, Sinan and Wang, Shijie and Fan, Zhihao and Bai, Jinze and Chen, Keqin and Liu, Xuejing and Wang, Jialin and Ge, Wenbin and others},
  journal={arXiv preprint arXiv:2409.12191},
  year={2024}
}

@inproceedings{babnik2022faceqan,
  title={{FaceQAN}: Face Image Quality Assessment Through Adversarial Noise Exploration},
  author={Babnik, {\v{Z}}iga and Peer, Peter and {\v{S}}truc, Vitomir},
  booktitle={2022 26th International Conference on Pattern Recognition (ICPR)},
  pages={748--754},
  year={2022},
  organization={IEEE}
}

@inproceedings{ou2021sddfiqa,
  title={{SDD-FIQA}: Unsupervised Face Image Quality Assessment with Similarity Distribution Distance},
  author={Ou, Fu-Zhao and Chen, Xingyu and Zhang, Ruixin and Huang, Yuge and Li, Shaoxin and Li, Jilin and Li, Yong and Cao, Liujuan and Wang, Yuan-Gen},
  booktitle={Proceedings of the IEEE/CVF Conference on Computer Vision and Pattern Recognition},
  pages={7670--7679},
  year={2021}
}

@inproceedings{atzori2025vitfiqa,
  title={{ViT-FIQA}: Assessing Face Image Quality using Vision Transformers},
  author={Atzori, Andrea and Boutros, Fadi and Damer, Naser},
  booktitle={Proceedings of the IEEE/CVF International Conference on Computer Vision Workshops},
  volume={1},
  number={2},
  pages={3},
  year={2025}
}

@article{zheng2025qavlm,
  title={{QA-VLM}: Providing human-interpretable quality assessment for wire-feed laser additive manufacturing parts with Vision Language Models},
  author={Zheng, Qiaojie and Zhang, Jiucai and Gockel, Joy and Wakin, Michael B and Brice, Craig and Zhang, Xiaoli},
  journal={Journal of Manufacturing Processes},
  volume={160},
  pages={611--623},
  year={2026},
  publisher={Elsevier}
}

@article{chen2025mgffdvlam,
  title={{MGFFD-VLM}: Multi-Granularity Prompt Learning for Face Forgery Detection with {VLM}},
  author={Chen, Tao and Zhang, Jingyi and others},
  journal={arXiv:2507.12232},
  year={2025}
}

@inproceedings{chen2024dslfiqa,
  title={{DSL-FIQA}: Assessing Facial Image Quality via Dual-Set Degradation Learning and Landmark-Guided Transformer},
  author={Chen, Wei-Ting and Krishnan, Gurunandan and Gao, Qiang and Kuo, Sy-Yen and Ma, Sizhou and Wang, Jian},
  booktitle={Proceedings of the IEEE/CVF Conference on Computer Vision and Pattern Recognition},
  pages={2931--2941},
  year={2024}
}

@inproceedings{jo2023ifqa,
  title={{IFQA}: Interpretable Face Quality Assessment},
  author={Jo, Byungho and Cho, Donghyeon and Park, In Kyu and Hong, Sungeun},
  booktitle={Proceedings of the IEEE/CVF Winter Conference on Applications of Computer Vision},
  pages={3444--3453},
  year={2023}
}

@article{team2025gemma,
  title={Gemma 3 technical report},
  author={Team, Gemma and Kamath, Aishwarya and Ferret, Johan and Pathak, Shreya and Vieillard, Nino and Merhej, Ramona and Perrin, Sarah and Matejovicova, Tatiana and Ram{\'e}, Alexandre and Rivi{\`e}re, Morgane and others},
  journal={arXiv preprint arXiv:2503.19786},
  year={2025}
}

@article{laurencon2023obelics,
  title={{OBELICS}: An Open Web-Scale Filtered Dataset of Interleaved Image-Text Documents},
  author={Lauren{\c{c}}on, Hugo and Saulnier, Lucile and Tronchon, L{\'e}o and Bekman, Stas and Singh, Amanpreet and Lozhkov, Anton and Wang, Thomas and Karamcheti, Siddharth and Rush, Alexander and Kiela, Douwe and others},
  journal={Advances in Neural Information Processing Systems},
  volume={36},
  pages={71683--71702},
  year={2023}
}

@inproceedings{dan2023transface,
  title={{TransFace}: Calibrating Transformer Training for Face Recognition from a Data-Centric Perspective},
  author={Dan, Jun and Liu, Yang and Xie, Haoyu and Deng, Jiankang and Xie, Haoran and Xie, Xuansong and Sun, Baigui},
  booktitle={Proceedings of the IEEE/CVF International Conference on Computer Vision},
  pages={20642--20653},
  year={2023}
}

@inproceedings{you2025lvface,
  title={{LVFace}: Progressive Cluster Optimization for Large Vision Models in Face Recognition},
  author={You, Jinghan and Li, Shanglin and Sun, Yuanrui and Wei, Jiangchuan and Guo, Mingyu and Feng, Chao and Ran, Jiao},
  booktitle={Proceedings of the IEEE/CVF International Conference on Computer Vision},
  pages={11840--11849},
  year={2025}
}

@article{abdin2024phi,
  title={Phi-4 technical report},
  author={Abdin, Marah and Aneja, Jyoti and Behl, Harkirat and Bubeck, S{\'e}bastien and Eldan, Ronen and Gunasekar, Suriya and Harrison, Michael and Hewett, Russell J and Javaheripi, Mojan and Kauffmann, Piero and others},
  journal={arXiv preprint arXiv:2412.08905},
  year={2024}
}

@inproceedings{saritacs2024analyzing,
  title={Analyzing the Effect of Combined Degradations on Face Recognition},
  author={Sarita{\c{s}}, Erdi and Ekenel, Haz{\i}m Kemal},
  booktitle={2024 IEEE 18th International Conference on Automatic Face and Gesture Recognition (FG) Workshops},
  pages={1--5},
  year={2024},
  organization={IEEE}
}

@article{lewis2020retrieval,
  title={Retrieval-Augmented Generation for Knowledge-Intensive {NLP} Tasks},
  author={Lewis, Patrick and Perez, Ethan and Piktus, Aleksandra and Petroni, Fabio and Karpukhin, Vladimir and Goyal, Naman and K{\"u}ttler, Heinrich and Lewis, Mike and Yih, Wen-tau and Rockt{\"a}schel, Tim and others},
  journal={Advances in Neural Information Processing Systems},
  volume={33},
  pages={9459--9474},
  year={2020}
}
}

\end{document}